\documentclass{article}

\usepackage{arxiv}

\usepackage[utf8]{inputenc} % allow utf-8 input
\usepackage[T1]{fontenc}    % use 8-bit T1 fonts
\usepackage{hyperref}       % hyperlinks
\usepackage{url}            % simple URL typesetting
\usepackage{booktabs}       % professional-quality tables
\usepackage{amsfonts}       % blackboard math symbols
\usepackage{nicefrac}       % compact symbols for 1/2, etc.
\usepackage{microtype}      % microtypography
\usepackage{lipsum}

\usepackage{bm}
\usepackage{amsmath}
\usepackage{amssymb}
\usepackage{amsfonts}
\usepackage{graphicx}
\def\vec#1{\mathbf{#1}}
\usepackage{algorithm}
\usepackage{algorithmic}
\usepackage{comment}

\DeclareMathOperator*{\argmin}{arg\,min}

\usepackage{mathtools}
\usepackage{proof}

\usepackage{natbib}

\usepackage{amsthm}
\theoremstyle{plain}
\newtheorem{theorem}{Theorem}[section]

\newtheorem{lemma}[theorem]{Lemma}

\theoremstyle{definition}

\newtheorem{assumption}[theorem]{Assumption}
\theoremstyle{remark}

\title{Data-driven Projection Generation for Efficiently Solving 
Heterogeneous Quadratic Programming Problems}

\author{
  Tomoharu Iwata\\
  Communication Science Laboratories, NTT, Inc.\\
  \And
  Futoshi Futami\\
  The University of Osaka / The University of Tokyo / RIKEN AIP
}
\date{}

\begin{document}
\maketitle

\begin{abstract}
We propose a data-driven framework for efficiently solving quadratic programming (QP) problems by reducing the number of variables in high-dimensional QPs using instance-specific projection. A graph neural network-based model is designed to generate projections tailored to each QP instance, enabling us to produce high-quality solutions even for previously unseen problems. The model is trained on heterogeneous QPs to minimize the expected objective value evaluated on the projected solutions. This is formulated as a bilevel optimization problem; the inner optimization solves the QP under a given projection using a QP solver, while the outer optimization updates the model parameters. We develop an efficient algorithm to solve this bilevel optimization problem, which computes parameter gradients without backpropagating through the solver. We provide a theoretical analysis of the generalization ability of solving QPs with projection matrices generated by neural networks. Experimental results demonstrate that our method produces high-quality feasible solutions with reduced computation time, outperforming existing methods.
\end{abstract}

\renewcommand{\thefootnote}{}
\footnotetext{This work has been submitted to the IEEE for possible publication. Copyright may be transferred without notice, after which this version may no longer be accessible.}

\section{Introduction}
Quadratic programming (QP) forms a fundamental class of optimization problems with a wide range of applications in fields such as machine learning~\cite{geweke1986exact,rodriguez2010quadratic,butler2023efficient,mantel1969restricted,zanghirati2003parallel}, finance~\cite{best2000quadratic}, and control~\cite{petersen2006constrained}. Although many QP solvers have been developed~\cite{andersen2015cvxopt,stellato2020osqp}, solving QPs with many variables remains computationally demanding.

In addition to solver improvements, random projection-based methods have emerged as a promising approach for reducing problem dimensionality~\cite{vu2019random,d2020random,liberti2021random,fuji2022convexification}. These approaches project high-dimensional QPs into lower-dimensional subspaces, resulting in smaller surrogate problems that can be solved more efficiently. Importantly, these methods are solver-agnostic and can leverage advances in existing solvers. Furthermore, the projected solutions remain feasible for the original problems. However, a significant drawback is that the projection matrices are generated at random, often leading to low-quality solutions due to the lack of instance-specific tailoring.

To address this limitation, we propose a data-driven projection approach that learns to generate instance-specific projections using a neural network. Our method trains the neural network using heterogeneous QPs to produce projections that yield high-quality solutions even for unseen problems. While data-driven projection methods have been successfully used for linear programming~\cite{sakaue2024generalization,iwata2025learning}, they are inapplicable to QPs. 
%In this work, we focus on convex QPs with linear inequality constraints.

Our approach leverages a graph neural network (GNN)~\cite{morris2019weisfeiler} to construct projection matrices conditioned on each QP instance.
Our model is capable of handling heterogeneous QPs, including instances with varying numbers of variables and constraints.
The model is trained to minimize the expected original objective value when solving the projected problem. We formulate this training as a bilevel optimization, where the inner optimization solves the QP under the current projection, and the outer optimization updates the GNN parameters. To efficiently compute gradients for training, we apply the envelope theorem, which allows us to differentiate through the QP solution without tracking the optimization trajectory of the QP solver.

Our contributions are summarized as follows:
1)~We propose the first data-driven projection framework for solving QPs.
2)~We develop a GNN-based model that generates instance-dependent projection matrices, along with its bilevel optimization-based training procedure.
3)~We theoretically analyze the generalization bound for solving QPs with projection matrices generated by neural networks.
4)~We provide extensive empirical results showing that our method efficiently produces high-quality solutions compared to existing baselines.

\section{Related work}

A variety of random projection-based techniques have been developed to reduce the dimensionality of mathematical programming problems, including not only QP but also linear programming and semidefinite programming~\cite{vu2019random,poirion2023random,d2020random,fuji2022convexification,akchen2024column}.
These methods apply randomly generated projection matrices to reduce the number of variables or constraints, enabling more efficient problem solving. However, due to the randomness of the projection, they often yield low-quality solutions, particularly when the projection does not align well with the structure of the problem.
To resolve this issue, data-driven projection methods have recently been proposed for linear programming~\cite{sakaue2024generalization,iwata2025learning}. These approaches train projection based on LP instances, leading to improved solution quality over random projections.
Recently, a theoretical analysis of data-driven projection for QP was presented~\cite{nguyen2025provably}.
However, this work is theoretical, does not propose 
a specific methodology for generating projections, 
and does not provide empirical validation.
In addition, our analysis offers several advantages as described 
in Section~\ref{sec:theory}.

Neural networks have been applied to directly predict solutions for QPs~\cite{bertsimas2022online,gao2021deep,sambharya2023end,tan2024ensemble,wang2020learning,chen2018approximating,karg2020efficient,liu2020revocable,nowak2018revised,pei2023reinforcement,wang2020combinatorial}.
These methods are typically trained by minimizing the discrepancy between predicted and optimal solutions, while enforcing constraint satisfaction. Such approaches fully rely on learning-based approximations and do not leverage algorithmic solvers.
In contrast, our method integrates machine learning with solvers; a neural network generates instance-specific projections, and the solution is computed via a QP solver. This hybrid strategy ensures both feasibility and high-quality performance, as demonstrated in our experiments, and allows us to benefit from the improved efficiency and reliability of modern QP solvers. 

Another stream of data-driven approaches for solving QPs is to use machine learning for accelerating the solving process~\cite{bonami2018learning,ichnowski2021accelerating,jung2022learning,getzelman2021learning,king2024metric}. These methods aim to improve solver efficiency by learning from prior runs or problem structure. Since our approach is solver-agnostic, it can be complementary to such acceleration methods, and potentially be combined with them for further performance gains.

\section{Proposed Method}

\begin{figure}
  \centering
  \includegraphics[width=23em]{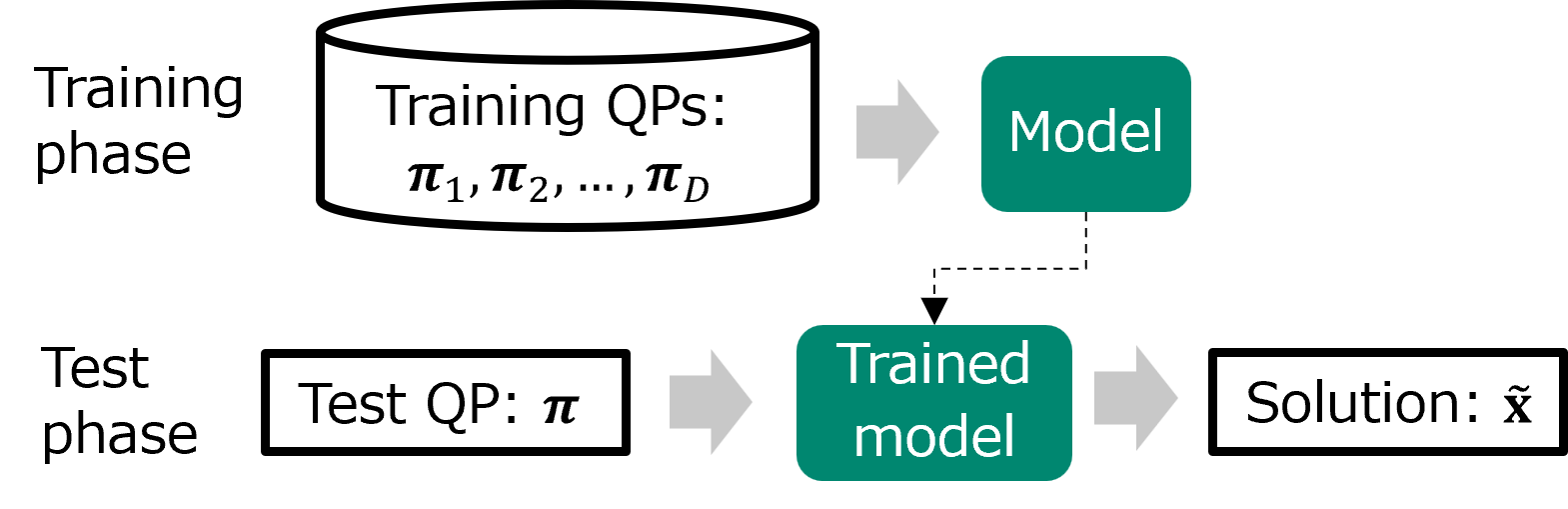}
  \caption{Our framework. In the training phase, our model is trained using multiple QPs. In the test phase, we obtain a solution of test QPs using the trained model, where training and test QPs are different.}
  \label{fig:framework}
\end{figure}

\begin{figure*}
  \centering
  \includegraphics[width=50em]{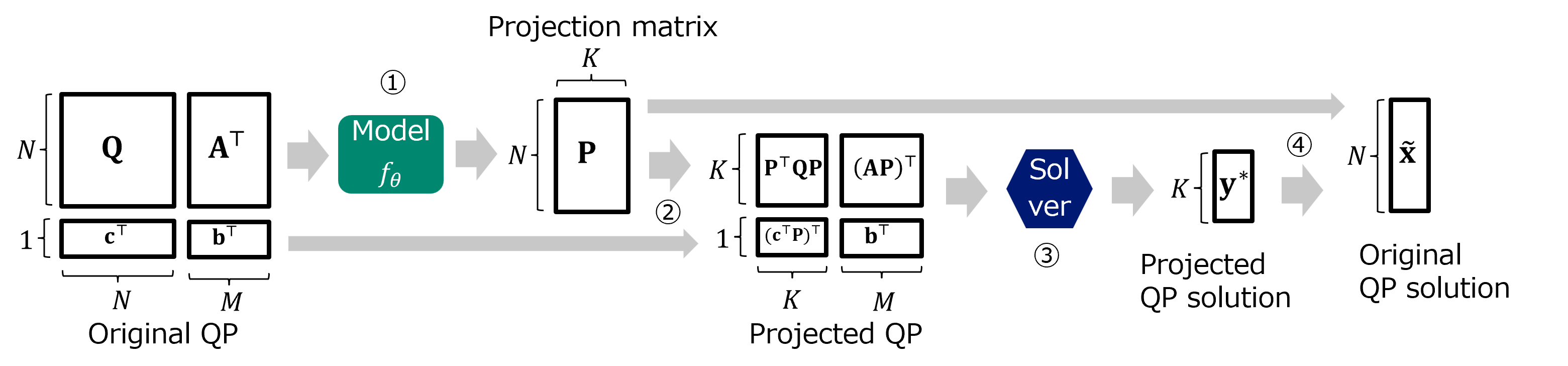}
  \caption{Our method to obtain a solution of a QP. 1) Given an original QP, an instance-specific projection matrix is generated using our model. Here, $N$ is the number of variables, $M$ is the number of constraints, and $K$ is the projection dimension. 2) The number of variables is reduced by the projection. 3) An optimal solution of the projected QP is obtained using a QP solver. 4) A solution of the original QP is recovered from the solution of the projected QP by $\tilde{\vec{x}} = \vec{P} \vec{y}^*$.}  
  \label{fig:method}
\end{figure*}

We propose a data-driven projection method for efficiently solving QPs.
An overview of our framework is presented in Figure~\ref{fig:framework}.
By generating training QPs that follow the parameter distribution of the target problems and pretraining our model on them, we can efficiently obtain high-quality feasible solutions for target QPs that will arise in the future.
The detailed steps of our method for solving a QP
are illustrated in Figure~\ref{fig:method}.

\subsection{Problem Formulation}

We consider quadratic programming (QP) problems with linear inequality constraints,
\begin{align}
  \underset{\vec{x} \in \mathbb{R}^{N}}{\text{min}} \quad
  \frac{1}{2} \vec{x}^\top \vec{Q} \vec{x} + \vec{c}^\top \vec{x}
  \quad \text{s.t.} \quad \vec{A} \vec{x} \leq \vec{b},
  \label{eq:qp}
\end{align}
where $\vec{Q} \in \mathbb{R}^{N \times N}$,
$\vec{c} \in \mathbb{R}^N$, $\vec{A} \in \mathbb{R}^{M \times N}$, and $\vec{b} \in \mathbb{R}^M$.  
This is a QP with $N$ variables and $M$ inequality constraints.  
Note that QPs with linear equality constraints can be equivalently transformed into this form without equality constraints, as described in Appendix~\ref{app:transform}.

In the training phase, we are given a dataset of $D$ QP instances, denoted by  
$\bm{\Pi} = \{ \bm{\pi}_d \}_{d=1}^{D}$,  
where each instance $\bm{\pi}_d = (\vec{Q}_d \in \mathbb{R}^{N_d \times N_d}, \vec{c}_d \in \mathbb{R}^{N_d}, \vec{A}_d \in \mathbb{R}^{M_d \times N_d}, \vec{b}_d \in \mathbb{R}^{M_d})$ represents a QP defined by its parameters.  
In the test phase,
we are given QP instances that are different from but related to the training QPs.
The number of variables and the number of constraints can be different across QP instances.
Our goal is to find high-quality solutions of the test QP instances efficiently.

\subsection{Dimensionality Reduction via Projection}

When the number of variables $N$ is large,
solving the original QP~\eqref{eq:qp} can be computationally expensive.
To improve efficiency, the solution can be approximated by restricting the optimization to a lower-dimensional subspace~\cite{d2020random}.
Specifically, a QP in Eq~(\ref{eq:qp}) is transformed using projection matrix $\vec{P}\in\mathbb{R}^{N\times K}$ with $K<N$,
\begin{align}
  \underset{\vec{y} \in \mathbb{R}^{K}}{\text{min}} \quad
  \frac{1}{2}
  \vec{y}^\top \vec{P}^\top \vec{Q} \vec{P} \vec{y} + \vec{c}^\top \vec{P} \vec{y}
  \quad \text{s.t.} \quad \vec{A} \vec{P} \vec{y} \leq \vec{b},
  \label{eq:projectedQP}
\end{align}
which defines a QP with $K$ variables and $M$ constraints.

Let $\vec{y}^*$ denote the optimal solution to the projected problem~\eqref{eq:projectedQP}.  
An approximate solution to the original QP is recovered
by mapping back via $\tilde{\vec{x}} = \vec{P} \vec{y}^*$.  
Since the constraint $\vec{A} \vec{P} \vec{y}^* \leq \vec{b}$ is satisfied in the projected space,  
the recovered solution $\tilde{\vec{x}}$ is guaranteed to be feasible
for the original QP, i.e., $\vec{A}\tilde{\vec{x}}\leq\vec{b}$, although it may not be optimal.
The quality of the recovered solution is evaluated using the original objective,
$\frac{1}{2} \tilde{\vec{x}}^\top \vec{Q} \tilde{\vec{x}} + \vec{c}^\top \tilde{\vec{x}} 
= \frac{1}{2} \vec{y}^{*\top} \vec{P}^\top \vec{Q} \vec{P} \vec{y}^* + \vec{c}^\top \vec{P} \vec{y}^* + \text{const}$.

\subsection{Projection Generation Models}

\begin{figure}
  \centering
  \includegraphics[width=16em]{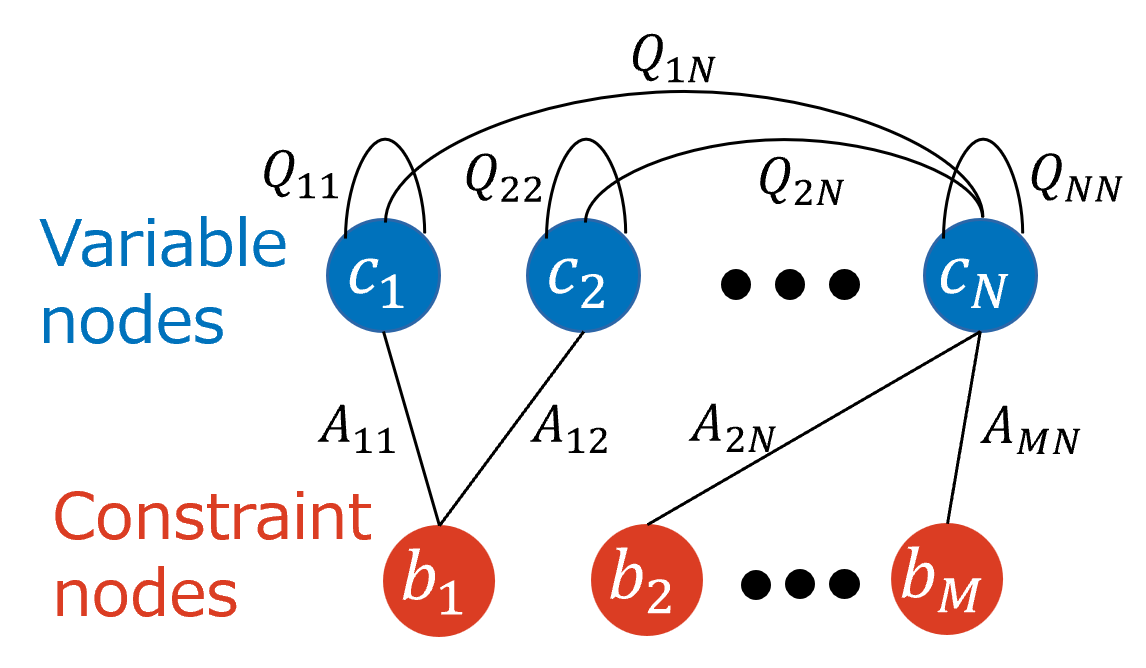}
  \caption{Graph for a QP. There are $N$ variable nodes and $M$ constraint nodes. Variable nodes are connected with weight $Q_{nn'}$. Variable and constraint nodes are connected with weight $A_{mn}$. Initial embeddings of variable and constraint nodes are calculated using QP parameters $c_{n}$ and $b_{m}$, respectively.}
  \label{fig:graph}
\end{figure}

We develop a GNN-based model, denoted by $f_{\bm{\theta}}(\bm{\pi})$, to generate an instance-specific projection matrix $\vec{P} \in \mathbb{R}^{N \times K}$ for a given QP instance $\bm{\pi} = (\vec{Q},\vec{c},\vec{A},\vec{b})$ with $N$ variables and $M$ constraints, where $\bm{\theta}$ represents the model parameters.
GNNs have been successfully used for QPs~\cite{chen2024expressive,wu2024representing}.
We construct a undirected graph consisting of $N$ variable nodes and $M$ constraint nodes using the QP parameters
as shown in Figure~\ref{fig:graph}.
An edge is placed between the $n$th and $n'$th variable nodes if $Q_{nn'} \neq 0$, with edge weight $Q_{nn'}$, which is the $(n,n')$ element of $\vec{Q}$.  
Additionally, an edge connects the $n$th variable node and the $m$th constraint node if $A_{mn} \neq 0$, with weight $A_{mn}$, which is the $(m,n)$ element of $\vec{A}$.  

The initial embedding for the $n$th variable node is computed by a linear transformation of $c_{n}$,
\begin{align}
  \vec{z}_{n0}^{\mathrm{V}} = \vec{w}_{0}^{\mathrm{V}} c_n + \vec{s}_{0}^{\mathrm{V}},
\end{align}
where $c_{n}\in\mathbb{R}$ is the $n$th element of $\vec{c}$.
Similarly, the initial embedding for the $m$th constraint node is computed by
\begin{align}
  \vec{z}_{m0}^{\mathrm{C}} = \vec{w}_{0}^{\mathrm{C}} b_m + \vec{s}_{0}^{\mathrm{C}},
\end{align}
where $b_{m}\in\mathbb{R}$ is the $m$th element of $\vec{b}$.
Here, $\vec{z}_{n\ell}^{\mathrm{V}}, \vec{z}_{m\ell}^{\mathrm{C}} \in \mathbb{R}^H$ represent the embedding vectors at the $\ell$th layer for the $n$th variable and $m$th constraint, respectively, vectors $\vec{w}_0^{\mathrm{V}}, \vec{w}_0^{\mathrm{C}} \in \mathbb{R}^H$ are learnable projection weights, and $\vec{s}_0^{\mathrm{V}}, \vec{s}_0^{\mathrm{C}} \in \mathbb{R}^H$ are bias terms.

At each GNN layer, the variable node embeddings are updated via message passing~\cite{morris2019weisfeiler},
\begin{align}
  \vec{z}_{n,\ell+1}^{\mathrm{V}} = \sigma \Biggl(
    \vec{W}_{\ell}^{\mathrm{V}} \vec{z}_{n\ell}^{\mathrm{V}} 
    + \frac{1}{|\mathcal{N}_{n}^{\mathrm{VV}}|} \vec{W}_{\ell}^{\mathrm{VV}} \sum_{n' \in \mathcal{N}_{n}^{\mathrm{VV}}} Q_{n'n} \vec{z}_{n'\ell}^{\mathrm{V}}
    + \frac{1}{|\mathcal{N}_{n}^{\mathrm{CV}}|} \vec{W}_{\ell}^{\mathrm{CV}} \sum_{m \in \mathcal{N}_{n}^{\mathrm{CV}}} A_{mn} \vec{z}_{m\ell}^{\mathrm{C}}
  \Biggr),
  \label{eq:message0}
\end{align}
where $\sigma$ is a nonlinear activation function, $\mathcal{N}_n^{\mathrm{VV}}$ and $\mathcal{N}_n^{\mathrm{CV}}$ denote the sets of neighboring variable and constraint nodes of the $n$th variable node, respectively, and $\vec{W}_{\ell}^{\mathrm{V}}, \vec{W}_{\ell}^{\mathrm{VV}}, \vec{W}_{\ell}^{\mathrm{CV}}\in\mathbb{R}^{H\times H}$ are linear projection matrices.
This update aggregates information from the node itself (first term), its neighboring variable nodes (second term), and its neighboring constraint nodes (third term), each transformed linearly.

The constraint node embeddings are similarly updated as,
\begin{align}
  \vec{z}_{m,\ell+1}^{\mathrm{C}} = \sigma \left(
    \vec{W}_{\ell}^{\mathrm{C}} \vec{z}_{m\ell}^{\mathrm{C}} 
    + \frac{1}{|\mathcal{N}_{m}^{\mathrm{VC}}|} \vec{W}_{\ell}^{\mathrm{VC}} \sum_{n \in \mathcal{N}_{m}^{\mathrm{VC}}} A_{mn} \vec{z}_{n\ell}^{\mathrm{V}}
  \right),
  \label{eq:message1}
\end{align}
where $\mathcal{N}_m^{\mathrm{VC}}$ denotes the variable neighbors of constraint node $m$.  
Eqs.~(\ref{eq:message0},\ref{eq:message1}) are applied iteratively for $L$ layers, enabling each node to incorporate structural information from the full QP instance.

After the GNN layers, we generate projection matrix $\vec{P} = [\vec{p}_1^\top, \dots, \vec{p}_N^\top]^\top \in \mathbb{R}^{N \times K}$
using the final variable node embeddings $\{\vec{z}_{nL}\}_{n=1}^{N}$, 
where each row vector $\vec{p}_n \in \mathbb{R}^K$ is computed as,
\begin{align}
  \vec{p}_n = g(\vec{z}_{nL}^{\mathrm{V}}),
  \label{eq:p}
\end{align}
where $g: \mathbb{R}^H \to \mathbb{R}^K$ is a shared feed-forward neural network applied identically to all variable nodes.
The outputs of $g$ are orthogonalized using QR decomposition 
to enforce linear independence among the columns of $\vec{P}$.

Parameters $\bm{\theta}$ in our model are initial embedding parameters
$\vec{w}_0^{\mathrm{V}}, \vec{s}_0^{\mathrm{V}}, \vec{w}_0^{\mathrm{C}}, \vec{s}_0^{\mathrm{C}}$,  
GNN weight matrices $\{ \vec{W}_\ell^{\mathrm{V}}, \vec{W}_\ell^{\mathrm{VV}}, \vec{W}_\ell^{\mathrm{CV}}, \vec{W}_\ell^{\mathrm{C}}, \vec{W}_\ell^{\mathrm{VC}} \}_{\ell=1}^L$,  
and the parameters of output network $g$.
All the model parameters are independent of the number of variables $N$ and constraints $M$, allowing our model to be applicable to QP instances of varying sizes.
Moreover, our model is permutation equivariant with respect to variables (i.e., permuting the variable order leads to a corresponding permutation in the projection rows), and permutation invariant with respect to constraints (i.e., permuting the constraints does not change the output).
These symmetry properties align with the invariance of QP solutions and reduce the effective hypothesis space of the model, enhancing generalization.

\subsection{Training}

We train our model by minimizing the expected objective value when the projection matrices generated by our model are applied to obtain solutions,
\begin{align}
  \hat{\bm{\theta}} = \underset{\bm{\theta}}{\argmin}
  \; \mathbb{E}_{\bm{\pi}_{d} \sim \bm{\Pi}} \left[ u(\vec{P}_{d}, \bm{\pi}_{d}) \right],
  \label{eq:theta}
\end{align}
where
$\mathbb{E}_{\bm{\pi}_{d}\sim\bm{\Pi}}$ is the expectation over the training QPs,
$\vec{P}_{d}=f_{\bm{\theta}}(\bm{\pi}_{d})$
is the projection matrix generated by our model,
and 
\begin{align}
  u(\vec{P}, \bm{\pi}) = 
  \min_{\vec{y} \in \mathbb{R}^{K}}
  \left\{
  \frac{1}{2} \vec{y}^\top \vec{P}^\top \vec{Q} \vec{P} \vec{y}
  + \vec{c}^\top \vec{P} \vec{y}
  \; \middle| \;
  \vec{A} \vec{P} \vec{y} \leq \vec{b}
  \right\}.
  \label{eq:u}
\end{align}
is the optimal objective value of the projected QP.
This is a bilevel optimization problem, where the inner optimization solves the projected QP and the outer optimization updates model parameters $\bm{\theta}$ to improve solution quality.

We solve this bilevel problem using stochastic gradient descent.  
The gradient of objective $u$ with respect to $\bm{\theta}$ is computed via the chain rule,
\begin{align}
  \frac{\partial u(\vec{P}_{d}, \bm{\pi}_{d})}{\partial \bm{\theta}}
  = 
  \mathrm{vec}\left(
    \frac{\partial u(\vec{P}_{d}, \bm{\pi}_{d})}{\partial \vec{P}_{d}}
  \right)^\top
  \frac{\partial \mathrm{vec}(\vec{P}_{d})}{\partial \bm{\theta}},
  \label{eq:dudtheta}
\end{align}
where $\mathrm{vec}(\cdot)$ denotes vectorization.  
The first term is computed using the envelope theorem~\cite{milgrom2002envelope},
\begin{align}
  \frac{\partial u(\vec{P}_{d}, \bm{\pi}_{d})}{\partial \vec{P}_{d}} 
  = \vec{Q}_{d} \vec{P}_{d} \vec{y}_{d}^{*} \vec{y}_{d}^{*\top}
  + \vec{c}_{d} \vec{y}_{d}^{*\top}
  + \vec{A}_{d}^\top \bm{\lambda}_{d}^{*} \vec{y}_{d}^{*\top},
\end{align}
where $\vec{y}^{*}$ is the optimal solution of the projected QP, and
$\bm{\lambda}^{*}$ is its dual solution.
Here, differentiation through the inner QP optimization is unnecessary, allowing the use of any QP solver that yields both primal and dual optimal solutions.
%The second term in Eq.~\eqref{eq:dudtheta} is computed via automatic differentiation.
Eq.~(\ref{eq:dudtheta}) can be efficiently evaluated via automatic differentiation by applying it to the scalar value
$\mathrm{vec}(\vec{Q}_{d}\vec{P}_{d}\vec{y}_{d}^{*}\vec{y}_{d}^{*\top}+\vec{c}_{d}\vec{y}_{d}^{*\top}+\vec{A}_{d}^{\top}\bm{\lambda}_{d}^{*}\vec{y}_{d}^{*\top})^{\top}\mathrm{vec}(\vec{P}_{d})$
with respect to model parameters $\bm{\theta}$,
where the first factor is frozen, excluding it from automatic differentiation.
This approach avoids the need to explicitly compute the potentially large Jacobian $\frac{\partial \mathrm{vec}(\vec{P}_{d})}{\partial \bm{\theta}}$.
Algorithm~\ref{alg:train} shows the training procedure.

\begin{algorithm}[t]
  \centering
  \caption{Training procedure of the proposed model.}
  \label{alg:train}
  \begin{algorithmic}[1]
    \REQUIRE{Training set $\bm{\Pi}$, batch size $B$}
    \ENSURE{Learned model parameters $\bm{\theta}$}
    \WHILE{termination criterion not met}
      \STATE Initialize total loss $\mathcal{L} \leftarrow 0$
      \STATE Sample minibatch $\bm{\Pi}_{\mathrm{B}} \subset \bm{\Pi}$ of $B$ QP instances
      \FOR{each QP $\bm{\pi} \in \bm{\Pi}_{\mathrm{B}}$}
        \STATE Generate projection matrix $\vec{P} \leftarrow f_{\bm{\theta}}(\bm{\pi})$
        \STATE Solve projected QP to obtain $\vec{y}^{*}$ (primal) and $\bm{\lambda}^{*}$ (dual) using a QP solver
        \STATE Update loss $\mathcal{L} \leftarrow \mathcal{L}+\mathrm{vec}(\vec{Q}\vec{P}\vec{y}^{*}\vec{y}^{*\top}+\vec{c}\vec{y}^{*\top}+\vec{A}^{\top}\bm{\lambda}^{*}\vec{y}^{*\top})^{\top}\mathrm{vec}(\vec{P})$
    \ENDFOR
    \STATE Calculate gradient of loss $\frac{\partial \mathcal{L}}{\partial\bm{\theta}}$ using automatic differentiation
    \STATE Update parameters $\bm{\theta}$ using the gradients
    by stochastic gradient descent
    \ENDWHILE
  \end{algorithmic}
\end{algorithm}

\section{Theoretical Analysis}
\label{sec:theory}
We analyze the generalization ability of an approach that solves QPs with projection matrices generated by neural networks. Let $\varPi$ be a set of QP instances with $N$ variables and $M$ constraints. Assume the optimal objective value of any QP in $\varPi$ lies in the interval $[-B,B]$. For every $\vec{P}\in\mathcal{P}$ and $\bm{\pi}=(\vec{Q},\vec{c},\vec{A},\vec{b})\in\varPi$, define $u(\vec{P},\bm{\pi})$ as in Eq.~(\ref{eq:u}). Let $f\in\mathcal{F}:\varPi\to\mathcal{P}$ be a neural network that maps a QP instance to a projection matrix. Then, $u\bigl(f(\bm{\pi}),\bm{\pi}\bigr)$ is the objective value of the QP associated with $\bm{\pi}$.

We theoretically analyze the generalization error, which measures the performance gap of $u(f(\cdot),\cdot)$ between the training and unseen test datasets. We assume a distribution $\mathcal{D}$ over $\varPi$ and an i.i.d.\ sample $\{\bm{\pi}_{d}\}_{d=1}^{D}\sim\mathcal{D}^{D}$ used to learn $f$. Under this setting, our goal is to evaluate the generalization error defined as $
\frac1D\sum_{d=1}^{D} u(f(\bm{\pi}_d),\bm{\pi}_d) - \mathbb{E}_{\bm{\pi}\sim\mathcal{D}}\bigl[u(f(\bm{\pi}),\bm{\pi})\bigr]$. We introduce technical assumptions for $\bm{\pi}$ and $f$ as follows. 
\begin{assumption}\label{asm_1}
   {\bf i)} For any $\bm{\pi}\in\varPi$, $\vec{Q}$ is positive definite with minimum eigenvalue $\sigma_{\mathrm{Q}}>0$ and satisfies $\|\vec{Q}\|_1<Q_0$, while the cost vector obeys $\|\vec{c}\|_1\le c_0$. {\bf ii)} For any $\bm{\pi}$ and any $f\in\mathcal{F}$, the projection matrix $\vec{P}=f(\bm{\pi})$ has minimum singular value at least $\sigma_{\mathrm{P}}>0$ and entries are in $[-1,+1]$.
\end{assumption}
These conditions exclude unbounded QPs; alternatively, one could assume that the feasible region of $\vec{y}$ is bounded (see Appendix~\ref{app_proof}).

We use covering numbers to evaluate the generalization gap. For $\bm{\pi}^D:=(\bm{\pi}_1,\dots,\bm{\pi}_D)\in\varPi^D$, define the pseudo‑metric $
d_D(f,g):= \max_{d\in[D]}\max_{j\in [NK]}\bigl|f(\bm{\pi}_d)_{j}-g(\bm{\pi}_d)_{j}\bigr|$, for $f,g\in\mathcal{F}$, where $f(\bm{\pi}_d)_{j}$ is the $j$th element of the $NK$-dimensional output $f(\bm{\pi}_d)$. The $\epsilon$-covering number of $\mathcal{F}$ with respect to $d_D$ is denoted by $\mathcal{N}(\epsilon,\mathcal{F},\bm{\pi}^D)$, and we define $\mathcal{N}(\epsilon,\mathcal{F},D):=\sup_{\bm{\pi}^D\in\varPi^D}\mathcal{N}(\epsilon,\mathcal{F},\bm{\pi}^D)$. Under these settings, the following result holds; its proof is deferred to Appendix~\ref{app_proof}.
\begin{theorem}\label{thm_generalization}
Under Assumption~\ref{asm_1}, for any $\varepsilon>0$ and $\delta\in(0,1)$, with probability at least $1-\delta$ over the draws of $\{\bm{\pi}_{d}\}_{d=1}^{D}$, we have
\begin{align}
\sup_{f\in\mathcal{F}}\Bigl|\,\frac1D\sum_{d=1}^{D} u(f(\bm{\pi}_d),\bm{\pi}_d) -\mathbb{E}_{\bm{\pi}\sim\mathcal{D}}\bigl[u(f(\bm{\pi}),\bm{\pi})\bigr]
\Bigr|  \notag
\leq  C(\varPi)\epsilon+\sqrt{\frac{8B^2\log\mathcal{N} (\epsilon,\mathcal{F},D)}{D}}+\sqrt{\frac{2B^2\log \frac{2}{\delta}}{D}},\notag
\end{align}
where $C(\varPi)$ depends only on $(\sigma_{\mathrm{Q}},\sigma_{\mathrm{P}},Q_0,c_0,B)$; its explicit form appears in Appendix~\ref{app_proof}.
\end{theorem}
\noindent\textbf{Implications.}  
Theorem~\ref{thm_generalization} guarantees that the generalization error decreases as $D$ grows and can be made arbitrarily small.  
Specifying $\mathcal{F}$ yields explicit rates via $\mathcal{N}(\epsilon,\mathcal{F},D)$.  
If $f\in\mathcal{F}$ is $\beta$-Lipschitz in its parameter $\bm{\theta}\in\mathbb{R}^{W}$, a standard covering argument~\cite{wainwright2019high} gives $\log\mathcal{N}(\epsilon,\mathcal{F},D) =\mathcal{O}(NKW\log\frac{\beta}{\epsilon})$. Setting $\epsilon=\mathcal{O}\!\bigl(1/(C(\varPi)\sqrt{D})\bigr)$ leads to the bound of Theorem~\ref{thm_generalization}
%\begin{align}
%    &\sup_{f\in\mathcal{F}}\Bigl|\,\frac1D\sum_{d=1}^{D} u(f(\pi_d),\pi_d) -\mathbb{E}_{\pi\sim\mathcal{D}}\bigl[u(f(\pi),\pi)\bigr]\Bigr|\notag \\
\begin{align}
%\leq \mathcal{O}\left(\sqrt{\frac{NKWB^2\log (D \beta C(\bm{\pi}))}{D}}\right)+\sqrt{\frac{2B^2\log\frac{2}{\delta}}{D}},\notag
\leq \mathcal{O}\left(\sqrt{\frac{NKWB^2\log (D \beta C(\varPi))}
{D}}\right)+\sqrt{\frac{2B^2\log\frac{2}{\delta}}{D}},\notag
\end{align}
which clarifies the convergence rate with respect to $D$. Appendix~\ref{app_proof} discusses more specific neural network architectures, which yield tighter bounds.

\paragraph{Comparison with existing work~\cite{nguyen2025provably}}
Here, we discuss the differences between our study and the recent existing work~\cite{nguyen2025provably}. 
While both studies analyze uniform convergence, they differ substantially in their \emph{assumptions} and \emph{analytical approaches}. 

Regarding the assumptions, both introduce the positive definiteness of the matrix $Q$ and the boundedness of the QP solution. 
The key difference is that existing work~\cite{nguyen2025provably} assumes a \emph{bounded feasible region} that contains the zero vector, 
whereas our analysis imposes assumptions on  \emph{cost vector} $\vec{c}$ and \emph{projection matrix} $\vec{P}$. 
Specifically, our assumptions allow us to bound the feasible solution region. 
In fact, by replacing our assumptions with those of existing work~\cite{nguyen2025provably}, the proof can be simplified without loss of generality. 

Next, we highlight the difference in proof techniques. 
In the existing work~\cite{nguyen2025provably}, the complexity term required for establishing uniform convergence is evaluated based on both the complexity of the QP problem itself and that of the neural network, using the \emph{pseudo-dimension}.  This approach can be regarded as an extension of the existing analysis (such as~\cite{sakaue2024generalization}) for LP problems to the QP setting. 
In contrast, our analysis focuses on the \emph{smoothness} property of the QP problem. 
We show that, under our assumptions, the QP solution is \emph{Lipschitz continuous} with respect to the neural network outputs. This allows us to evaluate the complexity term required for uniform convergence solely in terms of the neural network’s complexity, without analyzing the intrinsic complexity of the QP problem. 
Moreover, our approach is not limited to QP problems; it can be applied to a broader class of downstream optimization tasks that depend on neural network outputs, as long as the task exhibits sufficient smoothness. 
This represents a fundamental methodological difference from prior work.

%最後に既存研究\cite{nguyen2025provably}との差について，どちらも一様収束について述べている点で類似しているが仮定と解析方法の際について大きな差異が存在する．まず仮定について，どちらも$Q$の正定値性とQP解の有界性が導入されている．違う点は\cite{nguyen2025provably}らの研究ではfeasible regionがboundedで$0$ベクトルがそれに含まれるとしている．一方で本研究ではコストベクトルや射影ベクトルPに関する仮定があげられる．実際のところ，本研究ではコストベクトルや射影ベクトルPに関する仮定を入れることでFeasible regionとなる解$y$の範囲を求めている．そのため，我々の仮定は\cite{nguyen2025provably}らの仮定と入れ替えることができ，その場合証明はより単純化される．最後に証明方法の違いであるが，既存研究ではuniform convergenceの導出に必要なcomplexity termの評価をQP問題自体の複雑さとニューラルネットの複雑さの両方を基にして，Pseudo dimensionを用いて評価している．これは既存のLP問題の解析方針のQP問題への拡張になっている．一方で我々はQP問題の持つ滑らかさに注目している．我々はQP問題が仮定の下でLipschitz性をニューラルネットワークの出力に対して持つことを用い，uniform convergenceに必要なcomplexity termの評価を，QP問題の複雑さをもちいずにニューラルネットの複雑さのみを用いて評価している．また我々のアプローチは，QP問題に限らず，ニューラルネットの出力を用いるdown stream taskの最適問題の複雑さを評価することなく，その問題に滑らかさがあれば適用可能であるという点で既存研究とは大きく異なっている．

\section{Experiments}

\subsection{Data}

We evaluated our method using the following three types of QP datasets: Regression, Portfolio, and Control.
Regression dataset contains QP instances of
constrained linear regression~\cite{geweke1986exact},
where $N=500$ linear coefficients were optimized
with $M=50$ linear inequality constraints and non-negative constraints.
Portfolio dataset consists of portfolio optimization instances
that minimize the variance (i.e., risk)~\cite{best2000quadratic},
where $N=500$ portfolio weights were optimized
with budget and target expected return constraints, and no short-selling.
Control dataset comprises quadratic optimal control problems
with linear dynamics and box constraints~\cite{petersen2006constrained},
where state and control variables over time were optimized.
The number of total variables was $N=500$.
The equality constraints in Portfolio and Control datasets were
eliminated using feasible solutions
as described in Appendix~\ref{app:transform}.
For each dataset, we generated 120 training,
40 validation, and 40 test QP instances.
The details of the datasets were described in Appendix~\ref{app:data}.

\subsection{Compared Methods}
\label{sec:compared}

We compared our proposed method, denoted as \textsf{Ours}, with five baseline methods: \textsf{Rand}, \textsf{PCA}, \textsf{SharedP}, \textsf{Direct}, and \textsf{Full}.
\textsf{Rand} is a random projection-based method~\cite{d2020random} that reduces the dimensionality of the QP by uniform randomly selecting $K$ variables.
\textsf{PCA} is a data-driven projection method that constructs a shared projection matrix across all QP instances by applying principal component analysis (PCA) to the set of optimal solutions obtained from the training QPs~\cite{sakaue2024generalization}.
\textsf{SharedP} also learns a projection matrix shared across QPs, but does so by directly optimizing the expected objective value over the training set~\cite{sakaue2024generalization}. 
Note that both \textsf{PCA} and \textsf{SharedP} were originally developed for linear programming problems in \cite{sakaue2024generalization}. We extend these approaches to QP in our evaluation.
\textsf{Direct} is a data-driven approach that employs a graph neural network to directly predict the solution vector given a QP instance~\cite{chen2024expressive}. This method uses the same GNN architecture as \textsf{Ours}, but outputs a solution instead of a projection matrix.
\textsf{Full} refers to solving the original QP without any dimensionality reduction, and thus serves as an upper-bound benchmark.
Among these methods, \textsf{Ours}, \textsf{Rand}, \textsf{PCA}, and \textsf{SharedP} are projection-based methods, while \textsf{Ours}, \textsf{PCA}, \textsf{SharedP}, and \textsf{Direct} are data-driven methods.

\subsection{Settings}
\label{sec:settings}

Our method employs a graph neural network (GNN) with $L = 4$ layers, $H = 32$ hidden units per layer, and ReLU activation. The GNN is implemented using PyG~\cite{fey2019fast}. The output function $g$ is modeled as a three-layered fully-connected neural network with 32 hidden units and Leaky ReLU activations.
We optimized the model parameters using Adam~\cite{kingma2014adam} with a batch size of eight and a learning rate of $10^{-3}$.
The number of training epochs was capped at 500
and also treated as a hyperparameter.
PyTorch~\cite{paszke2019pytorch} was used for implementation.
All (projected) QPs were solved using the OSQP
solver~\cite{stellato2020osqp} via the
CVXPY interface~\cite{diamond2016cvxpy}.  
To tune hyperparameters, we minimized a validation loss composed of the sum of the relative errors on validation instances and a penalty term for infeasible solutions. The penalty term was computed as the failure rate (i.e., proportion of instances with no feasible solution) multiplied by $10^6$
on the training and validation instances.
For evaluating computational time for solving QPs,
we used a machine with
Intel Xeon CPU E5-2667 v4 3.20GHz CPU, 128GB memory,
and NVIDIA GeForce GTX1080Ti GPU.

\textsf{SharedP} and \textsf{Direct} were trained using the same training procedures as our method. In \textsf{Direct}, the neural network architecture matched that of our model, but was used to generate the solution vector directly rather than a projection matrix.
The training objective for \textsf{Direct} was the mean squared error between the predicted solution $\vec{x}_d$ and the optimal solution $\vec{x}_d^*$,
$\|\vec{x}_d^* - \vec{x}_d\|^2$.
To enforce feasibility, we added a constraint violation penalty term,
$\| \max(\vec{A}_d \vec{x}_d - \vec{b}_d, 0) \|$,
weighted by a regularization coefficient selected from $\{10^{-1}, 1, 10, 10^2\}$.

\subsection{Evaluation Measurements}

We evaluated solution quality using the following relative error metric on test QP instances, 
%$\frac{\hat{u} - u^*}{u_0 - u^*}$,    
\begin{align}
\text{Error}=\frac{\hat{u} - u^*}{u_0 - u^*},        
\end{align}
where $\hat{u}$ is the objective value of the obtained solution,  
$u^*$ is the optimal objective value computed by solving the original QP, 
and $u_0 = 0$ is the objective value of a trivial feasible solution
$\vec{x} = \vec{0}$.
For any method that failed to produce a feasible solution,
we assigned a relative error of one,
which corresponds to that of the trivial solution.  
Lower relative error indicates better solution quality.
To assess computational efficiency, we measured the total runtime
required to obtain a solution. 
This includes both the time for generating a projection (if applicable)
and the time for solving the (projected) QP.

\subsection{Results}

\begin{figure*}
  \centering
  \includegraphics[width=30em]{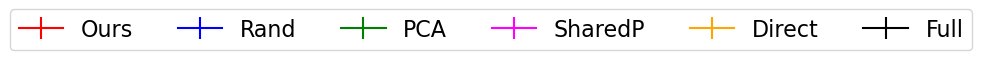}\\
  {\tabcolsep=0.3em\begin{tabular}{ccc}
  \includegraphics[width=15em]{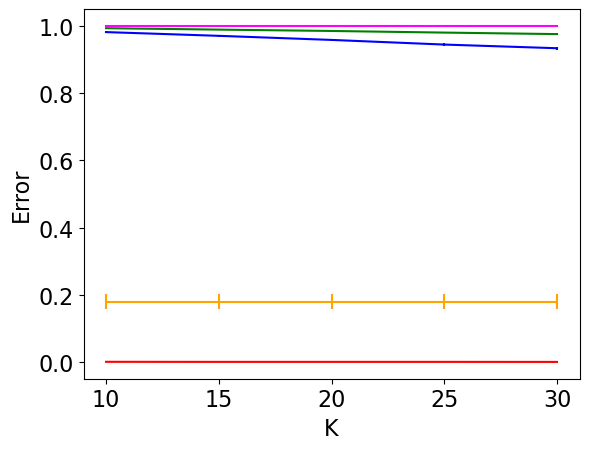}&
  \includegraphics[width=15em]{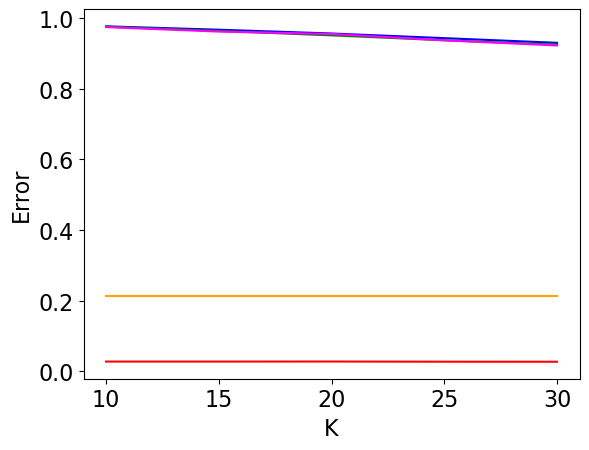}&
  \includegraphics[width=15em]{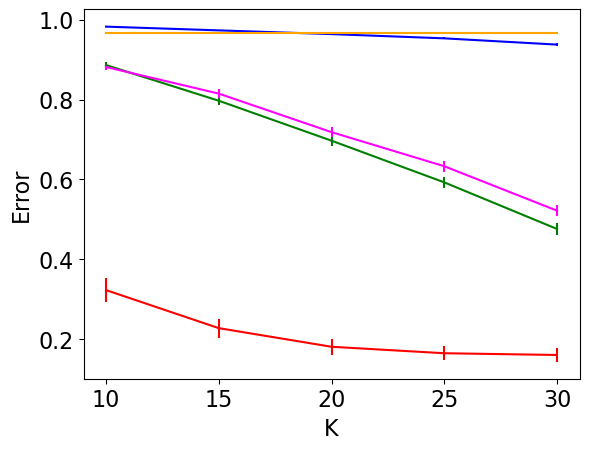}\\
  \includegraphics[width=15em]{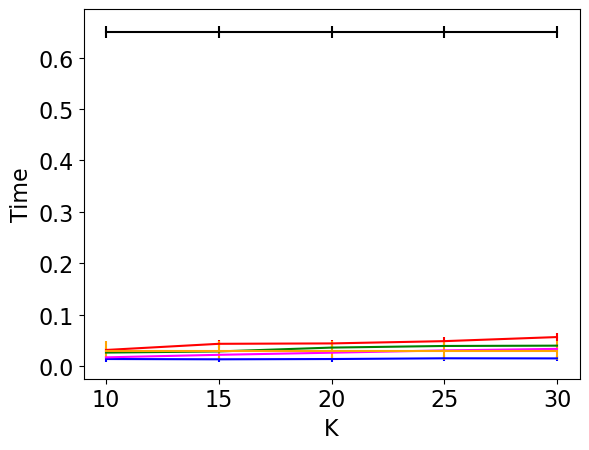}&
  \includegraphics[width=15em]{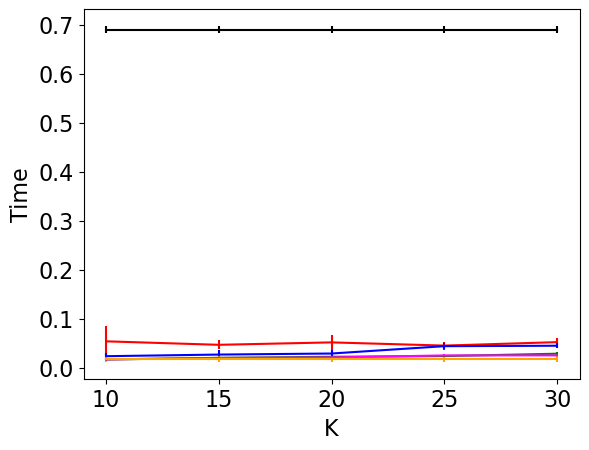}&
  \includegraphics[width=15em]{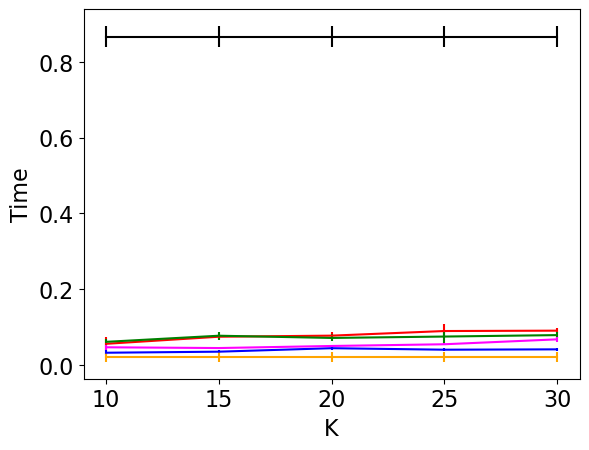}\\
  (a) Regression & (b) Portfolio & (c) Control \\
  \end{tabular}}
  \caption{Average relative errors (top) and computation times in seconds for solving a QP (bottom) with varying numbers of reduced variables $K$. Bars show standard errors.}
  \label{fig:error_k}
\end{figure*}

Figure~\ref{fig:error_k} presents the average relative errors and computation times for solving QPs using different numbers of reduced variables $K$. Our method consistently achieved low relative errors across all datasets.
For all projection-based methods, increasing the number of reduced variables generally led to improved accuracy. However, \textsf{Rand}, \textsf{PCA}, and \textsf{SharedP} performed poorly overall. The degradation in \textsf{Rand} is due to the use of random projection matrices, while both \textsf{PCA} and \textsf{SharedP} rely on a fixed projection matrix shared across all QP instances, which fails to capture instance-specific structure.
The error of \textsf{Direct} was higher than that of our method. This result suggests that solving QPs solely with a GNN-based predictor can be unreliable. In contrast, combining a GNN with QP solvers through our projection-based framework improved robustness across problem types.
In terms of feasibility, \textsf{Direct} failed to find feasible solutions for 5.8\% of test QPs, despite using regularization to penalize constraint violations. On the other hand, all projection-based methods, including ours, produced feasible solutions for all test cases. This highlights a key advantage of projection-based approaches; any solution to the projected QP is guaranteed to be feasible for the original QP, provided the projected problem is itself feasible and bounded. 

Regarding computational efficiency, our method was significantly faster than solving the full original QP. While slightly slower than other projection-based baselines due to the overhead of projection generation via a GNN, it still achieved substantial speedups. 
%For example, with our method, projection generation took 0.019 seconds and solving the projected QP required only 0.072 seconds on the Regression dataset with $K = 30$. 
  
\begin{figure*}[t!]
  \centering
  \includegraphics[width=30em]{images/legend.png}\\
  {\tabcolsep=0.3em\begin{tabular}{ccc}
  \includegraphics[width=15em]{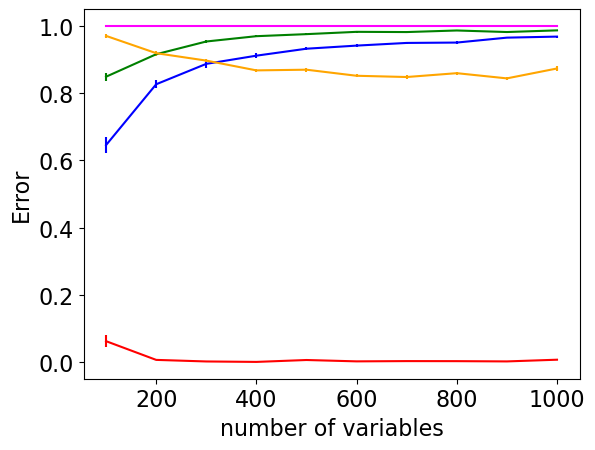}&
  \includegraphics[width=15em]{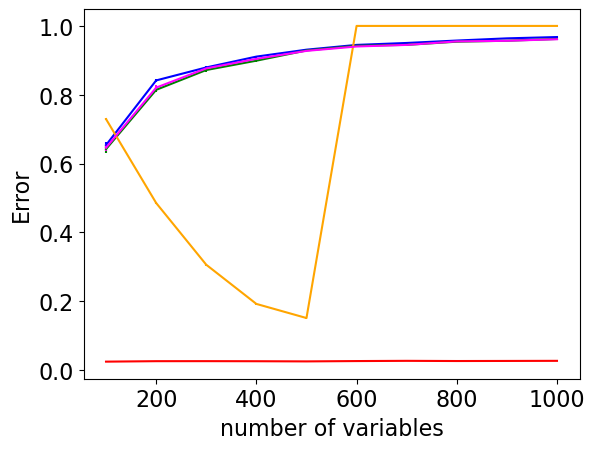}&
  \includegraphics[width=15em]{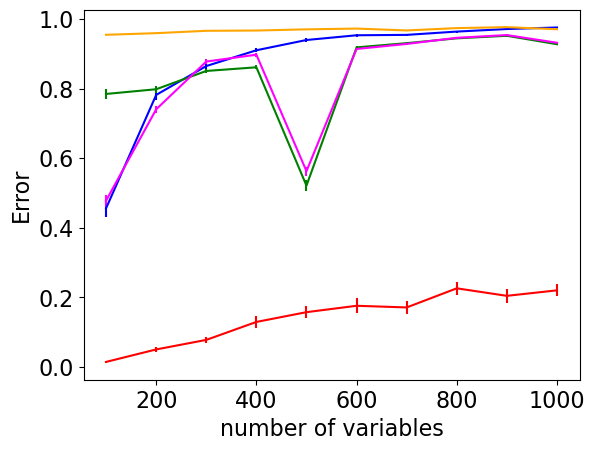}\\
  \includegraphics[width=15em]{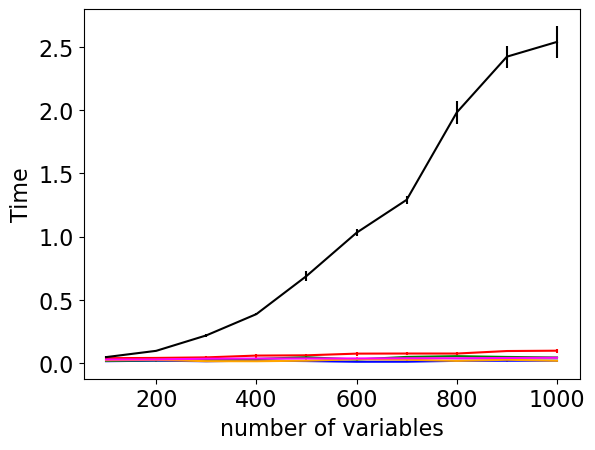}&
  \includegraphics[width=15em]{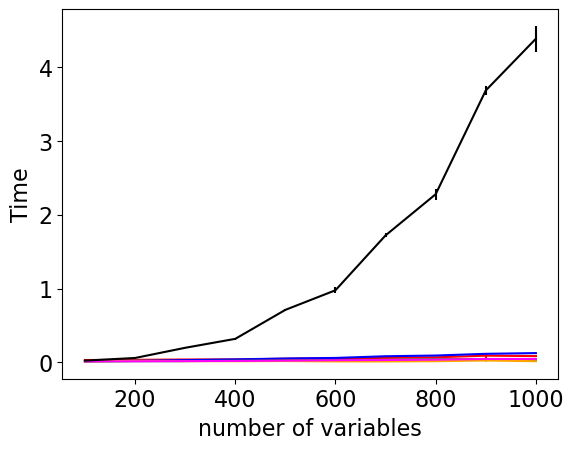}&
  \includegraphics[width=15em]{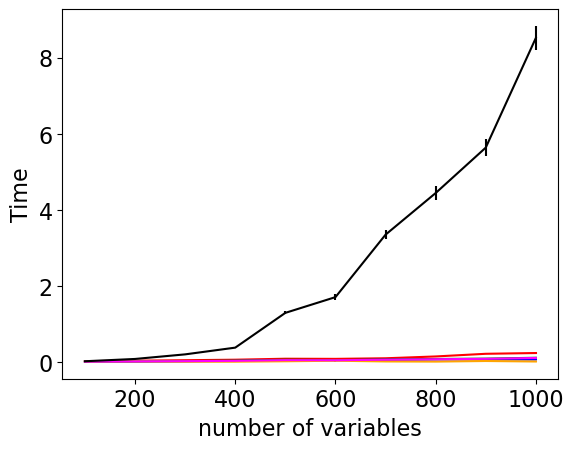}\\
  (a) Regression & (b) Portfolio & (c) Control \\
  \end{tabular}}
\caption{Average relative errors (top) and computation time in seconds for solving QPs (bottom) with varying numbers of variables $N$ in the test set. All data-driven methods were trained on QPs with $N = 500$, and projection-based methods used $K = 30$ reduced variables. Bars show the standard error.}
  \label{fig:error_n}  
\end{figure*}

Figure~\ref{fig:error_n} shows the average relative errors and computation times on test QPs with varying numbers of variables $N$. All data-driven methods were trained using QP instances with $N = 500$. In the Control dataset, we varied $N$ by adjusting the dimensionality of the state and control variables, setting $S = V$ to maintain balance.
Since \textsf{PCA} and \textsf{SharedP} use a single, fixed projection matrix across all QP instances, we padded their projection matrices with zeros when the test QPs contained more variables than seen during training. These methods generally exhibited poor generalization to variable-dimensional test cases.
In contrast, our method maintained low relative errors across different values of $N$, demonstrating robustness to changes in problem size. This suggests that our model effectively generalizes to QPs with unseen dimensionalities, thanks to our GNN-based model.
The error of \textsf{Rand} increased with $N$, as high-dimensional QPs are generally more difficult to approximate through random subspace selection. In some cases, \textsf{PCA}, \textsf{SharedP}, and \textsf{Direct} performed well only when $N = 500$, i.e., when the test QP size matched that of the training data. For example, \textsf{Direct} performed well on the Portfolio dataset when $N = 500$; the error of \textsf{PCA} and \textsf{SharedP} on the Control dataset dropped when $N=500$. These results highlight their limited adaptability to variable-sized QPs.
As expected, the computational time for solving full QPs increased substantially with $N$. In contrast, projection-based methods, including ours, remained efficient, showing only marginal increases in computation time regardless of the input dimension.

\begin{figure*}[t!]
  \centering
  \includegraphics[width=30em]{images/legend.png}\\  
  {\tabcolsep=0.3em\begin{tabular}{cc}
  \includegraphics[width=15em]{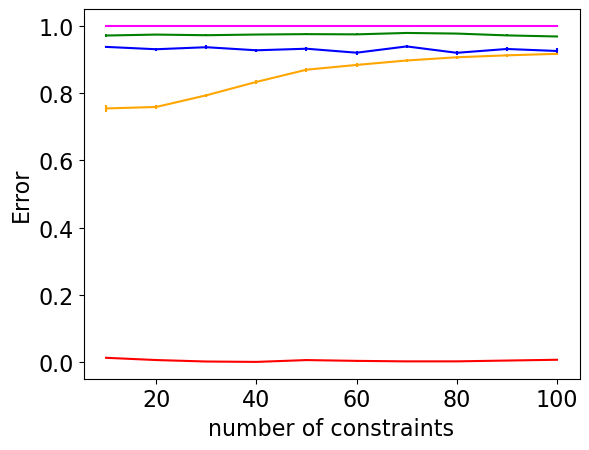}&
  \includegraphics[width=15em]{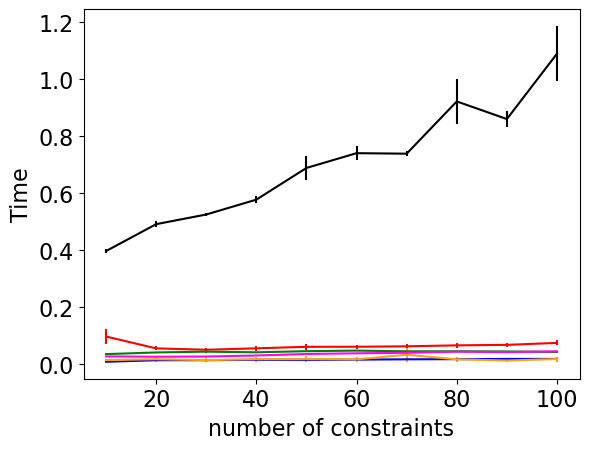}\\
  \end{tabular}}
\caption{Average relative errors (left) and computation time in seconds for solving QPs (right) with varying numbers of constraints $M$ in the test set on the Regression dataset. All data-driven methods were trained using QPs with $M = 50$, and projection-based methods used $K = 30$ reduced variables. Bars show the standard error.}
  \label{fig:error_m}    
\end{figure*}

Figure~\ref{fig:error_m} presents the average relative errors and computation time for solving test QPs with varying numbers of constraints $M$ on the Regression dataset. Among the three datasets, only Regression allows the number of constraints to be varied independently of the number of variables. Our method consistently achieved low error across different values of $M$, demonstrating robustness to changes in constraint dimensionality.

Table~\ref{tab:error_dataset} presents the average relative errors when our model was trained and tested on different datasets. For each case, the validation data matched the type of the training dataset. As expected, our method achieved the best performance when the training and test datasets were of the same type.

\begin{table*}[t!]
  \centering
\caption{Average relative errors and standard errors when training and test datasets differ. Bold values indicate the best results in each test dataset.}
  \label{tab:error_dataset}      
  \begin{tabular}{lrrr}
    \hline
    Train $\setminus$ Test & Regression & Portfolio & Control\\
    \hline
    Regression &
    {\bf 0.001 $\pm$ 0.000} &
    0.153 $\pm$ 0.012 &
    0.267 $\pm$ 0.152 \\
    Portfolio &
    0.613 $\pm$ 0.049 &
    {\bf 0.028 $\pm$ 0.003} &
    0.272 $\pm$ 0.172 \\
    Control &
    0.646 $\pm$ 0.039 &
    0.408 $\pm$ 0.021 &
    {\bf 0.154 $\pm$ 0.103} \\
    \hline
  \end{tabular}
  \end{table*}

\begin{figure*}[t!]
  \centering
  {\tabcolsep=0em\begin{tabular}{cc}
  \includegraphics[width=25em]{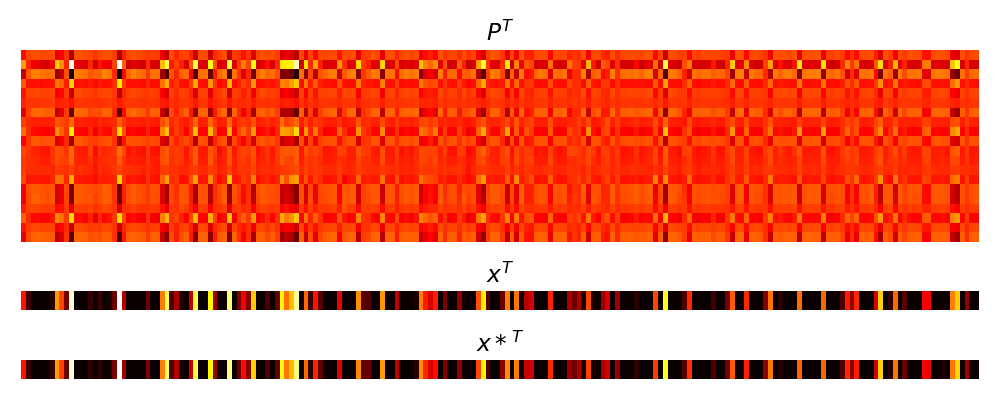}&
  \includegraphics[width=25em]{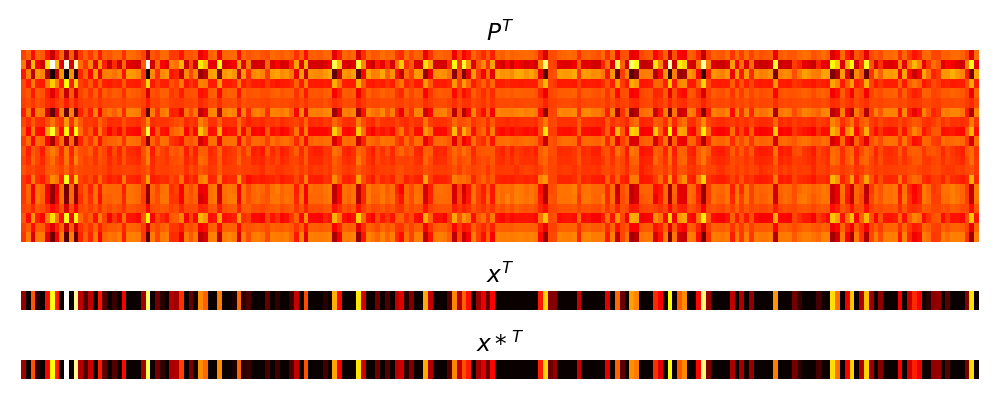}\\
  (a) QP1 & (b) QP2\\
  \end{tabular}}
  \caption{Examples of generated projection matrices (top) and corresponding solutions (middle) produced by our method with $K=20$ for two QP instances in the Regression dataset. The bottom row displays the optimal solutions. For clarity, 200 out of 500 variables are shown.}
  \label{fig:pxx}
\end{figure*}

Figure~\ref{fig:pxx} shows examples of generated projection and corresponding solutions by our method. The generated projection matrices are different across QP instances, by which we obtain solutions close to the optimal.

\begin{figure*}[t!]
  \centering
  \includegraphics[width=30em]{images/legend.png}\\    
  {\tabcolsep=0.3em\begin{tabular}{ccc}
  \includegraphics[width=15em]{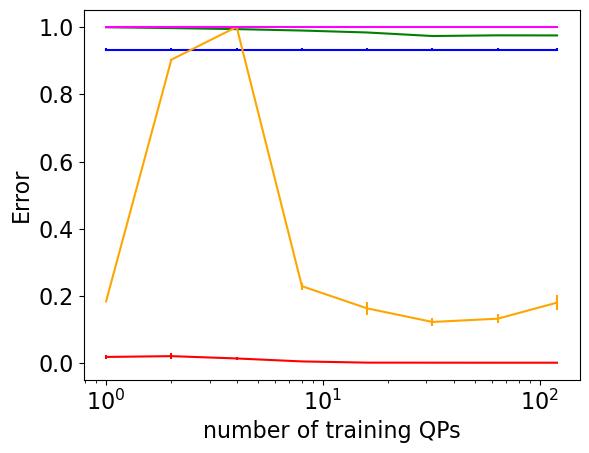}&
  \includegraphics[width=15em]{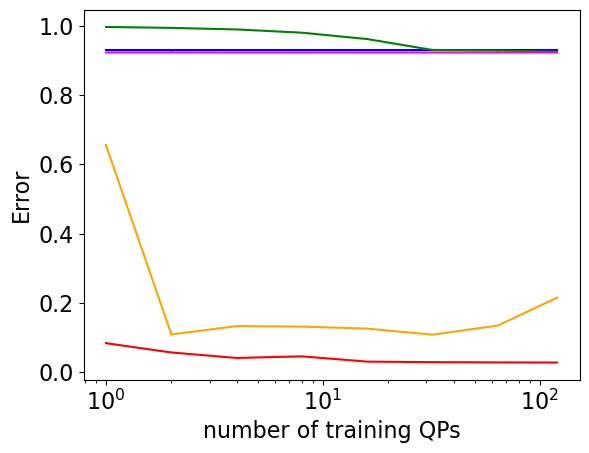}&
  \includegraphics[width=15em]{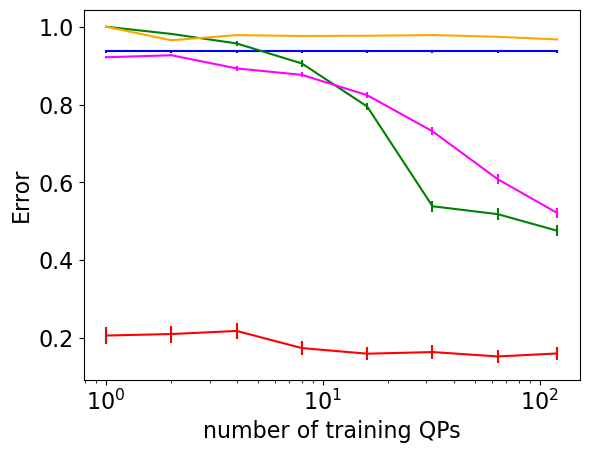}\\
  \end{tabular}}
  \caption{Average relative errors and standard errors with varying numbers of training QP instances. Bars show standard errors.}
  \label{fig:error_d}        
\end{figure*}

Figure~\ref{fig:error_d} shows the effect of the number of training QPs $D$ on the model's performance. As expected, relative errors decreased as more training instances were provided. Our method performed well even with a limited number of training samples, whereas the performance of \textsf{Direct} on the Regression dataset was unstable.

The training time required for our method with $K=30$ was 1.64 hours on the Regression dataset, 1.92 hours on the Portfolio dataset, and 2.58 hours on the Control dataset. While training our model incurs a computational cost, the trained model can produce solutions very efficiently at inference time. As shown in Figure~\ref{fig:error_k}, the average time to solve a test QP by our method was 0.054 seconds on Regression, 0.052 seconds on Portfolio, and 0.090 seconds on Control with $K=30$.

\section{Conclusion}

We proposed a data-driven method for efficiently solving QPs by reducing their size through instance-specific projections generated using a graph neural network.
We developed an efficient training procedure for our model.
We provided a theoretical analysis of the generalization ability for learning a neural network to generate projection matrices that reduce the size of QPs.
It guarantees that the generalization error decreases as the amount of training QPs increases, and can be made arbitrarily small.
Our experimental results show that our method produces high-quality feasible solutions, outperforming existing approaches while requiring significantly less computation time compared to solving the original QPs. Furthermore, it demonstrates robust generalization across QPs of varying sizes.
For future work, we plan to extend our method to a broader class of problems, including QPs with quadratic constraints. 
%In addition, we 
%While we believe our approach represents an important step toward more efficient QP solving, there are several directions for future work.
%First, we plan to extend our method to a broader class of problems, including nonconvex QPs and those with quadratic constraints. Second, we would like to estimate the distribution of training QPs.

\appendix

\section{Elimination of Equality Constraints}
\label{app:transform}

Consider a QP with both linear inequality and equality constraints,
\begin{align}
  \min_{\vec{u}\in\mathbb{R}^{N}} \frac{1}{2}\vec{u}^{\top}\vec{Q}'\vec{u}+\vec{c}'^{\top}\vec{u},
  \quad
  \text{s.t.}
  \quad
  \vec{A}_{\mathrm{I}}\vec{u}\leq\vec{b}_{\mathrm{I}},
  \quad
  \vec{A}_{\mathrm{E}}\vec{u}=\vec{b}_{\mathrm{E}},
  \label{eq:qp_eq}
\end{align}
where $\vec{Q}'\in\mathbb{R}^{N\times N}$,
$\vec{c}'\in\mathbb{R}^{N}$,
$\vec{A}_{\mathrm{I}}\in\mathbb{R}^{M_{\mathrm{I}}\times N}$,
$\vec{b}_{\mathrm{I}}\in\mathbb{R}^{M_{\mathrm{I}}}$
$\vec{A}_{\mathrm{E}}\in\mathbb{R}^{M_{\mathrm{E}}\times N}$, and
$\vec{b}_{\mathrm{E}}\in\mathbb{R}^{M_{\mathrm{E}}}$.
One simple way to eliminate equality constraints is to rewrite them as pairs of inequality constraints,
\begin{align}
  &\min_{\vec{u}\in\mathbb{R}^{N}} \frac{1}{2}\vec{u}^{\top}\vec{Q}'\vec{u}+\vec{c}'^{\top}\vec{u},
  \nonumber\\
  &\text{s.t.}
  \quad
  %\vec{A}\vec{q}\leq\vec{b},
  \vec{A}_{\mathrm{I}}\vec{u}\leq\vec{b}_{\mathrm{I}},
  \quad
  \vec{A}_{\mathrm{E}}\vec{u}\leq\vec{b}_{\mathrm{E}},
  \quad
  -\vec{A}_{\mathrm{E}}\vec{u}\leq-\vec{b}_{\mathrm{E}}.  
  \label{eq:qp_ineq1}
\end{align}
This is an inequality-form QP in Eq~(\ref{eq:qp}) with parameters defined by
$\vec{A}=[\vec{A}_{\mathrm{I}},\vec{A}_{\mathrm{E}},-\vec{A}_{\mathrm{E}}]\in\mathbb{R}^{(M_{\mathrm{I}}+2M_{\mathrm{E}})\times N}$,
and
$\vec{b}=[\vec{b}_{\mathrm{I}},\vec{b}_{\mathrm{E}},-\vec{b}_{\mathrm{E}}]\in\mathbb{R}^{M_{\mathrm{I}}+2M_{\mathrm{E}}}$.

Alternatively, when a (trivially) feasible solution $\vec{u}_{0}$ satisfying both
$\vec{A}_{\mathrm{I}}\vec{u}_{0}\leq\vec{b}_{\mathrm{I}}$ and
$\vec{A}_{\mathrm{E}}\vec{u}_{0}=\vec{b}_{\mathrm{E}}$ is available,
a more efficient transformation can be applied that preserves the original number of inequality constraints.
Letting $\vec{u} = \vec{u}' + \vec{u}_{0}$ shifts the variable around a known feasible point. Substituting into Eq.~(\ref{eq:qp_eq}) gives,
\begin{align}
  \min_{\vec{u}'\in\mathbb{R}^{N}} \frac{1}{2}(\vec{u}'+\vec{u}_{0})^{\top}\vec{Q}'(\vec{u}'+\vec{u}_{0})+
  \vec{c}'^{\top}(\vec{u}'+\vec{u}_{0}),
  \nonumber\\
  \text{s.t.}
  \quad
  \vec{A}_{\mathrm{I}}\vec{u}'\leq\vec{b}_{\mathrm{I}}-\vec{A}_{\mathrm{I}}\vec{u}_{0},
  \quad
  \vec{A}_{\mathrm{E}}\vec{u}'=\vec{0}.
  \label{eq:qp_eq2}
\end{align}
The equality constraint $\vec{A}_{\mathrm{E}}\vec{u}'=0$ implies that $\vec{u}'$ lies in the null space of $\vec{A}_{\mathrm{E}}$.
Let $\vec{D} = \vec{I} - \vec{A}_{\mathrm{E}}^{+} \vec{A}_{\mathrm{E}}$ be the orthogonal projection onto the null space of $\vec{A}_{\mathrm{E}}$,
where $\vec{A}_{\mathrm{E}}^{+}$ is the Moore–Penrose pseudo-inverse.
Then, $\vec{u}'$ can be expressed as $\vec{u}' = \vec{D}\vec{x}$ for some $\vec{x} \in \mathbb{R}^N$.
Substituting into Eq.~(\ref{eq:qp_eq2}) gives,
\begin{align}
  \min_{\vec{u}\in\mathbb{R}^{N}} \frac{1}{2}\vec{x}^{\top}\vec{D}^{\top}\vec{Q}'\vec{D}\vec{x}
  +(\vec{c}'^{\top}+\vec{u}_{0}^{\top}\vec{Q})\vec{D}\vec{x}
  \nonumber\\
  \text{s.t.}
  \quad
  \vec{A}_{\mathrm{I}}\vec{D}\vec{x}\leq\vec{b}_{\mathrm{I}}-\vec{A}_{\mathrm{I}}\vec{u}_{0},
  \label{eq:qp_eq3}
\end{align}
where constant term
$\frac{1}{2}\vec{u}_{0}^{\top}\vec{Q}'\vec{u}_{0}+\vec{c}'^{\top}\vec{u}_{0}$ is omitted.
The resulting formulation is an inequality-form QP in Eq~(\ref{eq:qp}) with parameters defined by
$\vec{Q}=\vec{D}^{\top}\vec{Q}'\vec{D}$,
$\vec{c}=(\vec{c}'^{\top}+\vec{u}_{0}^{\top}\vec{Q})\vec{D}$,
$\vec{A}=\vec{A}_{\mathrm{I}}\vec{D}$, and
$\vec{b}=\vec{b}_{\mathrm{I}}-\vec{A}_{\mathrm{I}}\vec{u}_{0}$.
For this transformed QP, $\vec{x}=0$ is always feasible.

\section{Proof of Theorem~\ref{thm_generalization} and additional discussion}
\label{app_proof}

\subsection{Proof of Theorem~\ref{thm_generalization}}
Following the standard approach in uniform convergence analysis~\cite{Mohri}, we derive the convergence in expectation as follows:

For any $N\in \mathbb{N}$, let $[N]=\{1,\dots,N\}$. Under the assumption that QPs are bounded, we have
\begin{align}
&\sup_{f \in \mathcal{F}} |\mathbb{E}_{\bm{\pi}\sim\mathcal{D}}u(f(\bm{\pi}),\bm{\pi}) - \frac1D\sum_{d=1}^{D} u(f(\bm{\pi}_d),\bm{\pi}_d)|\notag \\
&\leq \mathbb{E}_{\bm{\pi}^D \sim \mathcal{D}^{D}} \sup_{f \in \mathcal{F}} |\mathbb{E}_{\bm{\pi}'\sim\mathcal{D}}u(f(\bm{\pi}'),\bm{\pi}')-\frac1D\sum_{d=1}^{D} u(f(\bm{\pi}_d),\bm{\pi}_d)| \notag \\
&\quad+ \sqrt{\frac{2B^2\log\frac{2}{\delta}}{D}},
\end{align}
where this is the consequence of the McDiarmid's inequality~\cite{Mohri}. And then we apply the symmetrization argument,
\begin{align}
&\mathbb{E}_{(\bm{\pi}^D, \bm{\pi'}^D) \sim \mathcal{D}^{2D}} \notag \\
&\quad\quad\mathbb{E}_{\sigma} \sup_{f \in \mathcal{F}} \frac{1}{D} \sum_{d=1}^{D} 
\left[\sigma_d u(f(\bm{\pi'}_d),\bm{\pi'}_d) - \sigma_d u(f(\bm{\pi}_d),\bm{\pi}_d)\right]\notag\\
&\leq2\mathbb{E}_{\bm{\pi}^D \sim \mathcal{D}^{D}} \mathbb{E}_{\sigma} \sup_{f \in \mathcal{F}} \frac{1}{D} \sum_{d=1}^{D} 
\sigma_d u(f(\bm{\pi}_d),\bm{\pi}_d)\notag
\end{align}
where $\sigma=(\sigma_1,\dots,\sigma_D)$ are i.i.d. Rademacher variables. We define $R(\mathcal{F},D):=\mathbb{E}_{\sigma} \sup_{f \in \mathcal{F}} \frac{1}{D} \sum_{d=1}^{D} 
\sigma_d u(f(\bm{\pi}_d),\bm{\pi}_d)$

We now derive an upper bound using a variant of the Massart lemma. For simplicity, assume the function class \( \mathcal{F} \) has finite cardinality \( |\mathcal{F}| \). A covering number will later replace this. Then for all \( \lambda > 0 \), we have:
\begin{align}\label{eq_ref_massart}
&R(\mathcal{F}, D)\notag\notag \\
&\leq \mathbb{E}_{\sigma} \frac{1}{\lambda} \log \sum_{f \in \mathcal{F}} \exp \left(  \frac{\lambda}{D} \sum_{d=1}^{D} 
\sigma_d u(f(\bm{\pi}_d),\bm{\pi}_d)\right) \notag\\
&\leq \frac{1}{\lambda} \log \sum_{f \in \mathcal{F}} \mathbb{E}_{\sigma} \exp \left( \frac{\lambda}{D} \sum_{d=1}^{D} 
\sigma_d u(f(\bm{\pi}_d),\bm{\pi}_d) \right) \notag\\
&\leq \frac{1}{\lambda} \log |\mathcal{F}| \cdot \mathbb{E}_{\sigma} \exp \left(  \frac{\lambda}{D} \sum_{d=1}^{D} 
\sigma_d u(f(\bm{\pi}_d),\bm{\pi}_d) \right) \notag\\
&\leq \frac{1}{\lambda} \log |\mathcal{F}| e^{\frac{D}{8}(\lambda\frac{2B}{D})^2}= \frac{1}{\lambda} \log |\mathcal{F}|+ \frac{\lambda B^2}{2D}\notag\\
&\to \sqrt{\frac{2 B^2 \log |\mathcal{F}|}{D}}
\end{align}
where we optimized $\lambda$ in the last line.

So by the definition of the \( \epsilon \)-covering, for any \( f \in \mathcal{F} \), there exists \( \tilde f \in \mathcal{N}(\epsilon, \mathcal{F}, D) \) such that
\[
\max_{d\in[D]}\max_{j\in [NK]}\bigl|f(\bm{\pi}_d)_{j}-\tilde{f}(\bm{\pi}_d)_{j}\bigr| \leq \epsilon
%\|f - \tilde f\|_{L_2(S_{D})} \leq \epsilon.
\]

Next, we analyze how the discretization of $\mathcal{F}$ affects $u(f(\cdot),\cdot)$. To analyze this we focus on the Lipschitz continuity of $u$ as follows.
\begin{align}\label{eq_max_norm}
    &\frac{1}{D}\sum_{d=1}^D|u(f(\bm{\pi}_d),\bm{\pi}_d)-u(\tilde{f}(\bm{\pi}_d),\bm{\pi}_d)|\notag \\
    &\le C(\varPi) \max_{d\in[D]}\max_{j\in [NK]}\bigl|f(\bm{\pi}_d)_{j}-\tilde f(\bm{\pi}_d)_{j}\bigr|
    %&\leq \beta\frac{1}{D}\sum_{d=1}^D\sum_{j=1}^{NK}\bigl|f(\bm{\pi}_d)_{j}-g(\bm{\pi}_d)_{j}\bigr|^2=\beta\|f - \tilde f\|_{L_2(S_{D})} 
\end{align}
where $C(\varPi)$ is the Lipschitz constant, which is estimated in Appendix~\ref{app_lipshitz}. 

Then from the discretization argument, we have
\begin{align}
&\sup_{f \in \mathcal{F}} |\mathbb{E}_{\bm{\pi}\sim\mathcal{D}}u(f(\bm{\pi}),\bm{\pi}) - \frac1D\sum_{d=1}^{D} u(f(\bm{\pi}_d),\bm{\pi}_d)|\notag \\
&\leq  C(\varPi)\epsilon+2\sqrt{\frac{2 B^2 \log \mathcal{N}(\epsilon, \mathcal{F}, D)}{D}}+ \sqrt{\frac{2B^2\log\frac{2}{\delta}}{D}},
\end{align}
where we substitute $\mathcal{N}(\epsilon, \mathcal{F}, D)$ to $|\mathcal{F}|$.

\subsection{Estimate of the Lipschitz coefficient}\label{app_lipshitz}
As described above, we need to control the Lipschitzness of the discretization above.

\begin{lemma}[Lipschitz continuity of $u$ w.r.t.\ $\vec{P}$]
\label{lem:lipschitz_u}
Fix a QP instance $\bm{\pi}=(\vec{Q},\vec{c},\vec{A},\vec{b})$ that satisfies Assumption 1. For any two projection matrices $\vec{P},\vec{P}'\in[-1,1]^{N\times K}$ with
$\sigma_{\min}(\vec{P}),\sigma_{\min}(\vec{P}')\ge \sigma_{\mathrm{P}}$ we have
\[
    |u(\vec{P},\bm{\pi})-u(\vec{P}',\bm{\pi})|
    \;\le\;
    C(\varPi)'\,\|\vec{P}-\vec{P}'\|_{\infty},
\]
where
\[
    C(\varPi)'
    \;=\;
    Q_0\mu_{\mathrm{P}}\,Y_{\max}^{2}
    \;+\;
    c_{0}\,Y_{\max},
\]
and $\mu_{\mathrm{P}}:=NK$ and
\[
    Y_{\max}
    :=
    \sqrt{N}\frac{c_{0}
          +\sqrt{c_{0}^{2}+2\sigma_{\mathrm{Q}}B}}
         {\sigma_{\mathrm{Q}}}.
\]
\end{lemma}
Based on this lemma, the Lipshitz constant in max norm of Eq.~\eqref{eq_max_norm} is $C(\varPi)=\sqrt{NK}\beta'$

\begin{proof}
Here we first derive a Lipschitz constant in the $\ell_{1}\times$ Frobenius setting.
We first convert the difference into an inner product using the mean-value theorem.
For any $\vec{P},\vec{P}'\in\mathbb R^{N\times K}$ there exists $\tau\in[0,1]$ such that
        \[
          u(\vec{P},\pi)-u(\vec{P}',\pi)
           = 
          \bigl\langle
            \nabla_{\vec{P}}u \bigl((1-\tau)\vec{P}+\tau \vec{P}',\pi\bigr), 
            \vec{P}-\vec{P}'
          \bigr\rangle,
        \]
We then apply Hölder with the $(1,\infty)$ pair, that is, 
        \[
          \bigl|\langle A,B\rangle\bigr|
           \le 
          \|A\|_{1}\,\|B\|_{\infty}
           \le 
          \|A\|_{1}\,\|B\|_{F},
        \]
        because $\|B\|_{\infty}\le\|B\|_{F}$.

\begin{align}
             &|u(\vec{P},\pi)-u(\vec{P}',\pi)|\notag \\
             &\le 
            \Bigl( 
              \sup_{\tau\in[0,1]}
              \bigl\|\nabla_{\vec{P}}u \bigl((1-\tau)\vec{P}+\tau \vec{P}',\pi\bigr)\bigr\|_{1}
            \Bigr)
            \,
            \|\vec{P}-\vec{P}'\|_{F}   
\end{align}             
        Define the \emph{$\ell_{1}$-Lipschitz constant}
        \[
          C(\varPi):=
            \sup_{\tau\in[0,1]}
            \bigl\|
              \nabla_{\vec{P}}u \bigl((1-\tau)\vec{P}+\tau \vec{P}',\pi\bigr)
            \bigr\|_{1},
        \]
        so that $|u(\vec{P},\pi)-u(\vec{P}',\pi)|\le C(\varPi)\,\|\vec{P}-\vec{P}'\|_{F}$.
Next we upper bound this coefficient. From KKT analysis,
        $\nabla_{\vec{P}}u(\vec{P},\pi)=\vec{Q}\vec{P}\,\vec{y}^*\vec{y}^{*\top}+\vec{c}\,\vec{y}^{*\top}$, where $\vec{y}^*$ is a feasible solution.  Then
        \[
          \bigl\|\nabla_{\vec{P}}u(\vec{P},\pi)\bigr\|_{1}
           \le 
          \|\vec{Q}\|_{1}\,\|\vec{P}\|_{1}\,\|\vec{y}\|_{1}^{2}
          +\|\vec{c}\|_{1}\,\|\vec{y}\|_{1}.
        \]
Because every entry of \({\vec{P}}\) lies in \([-1,1]\), 
\(\|\tilde{\vec{P}}\|_{1}\le NK=\mu_{\mathrm{P}}\).

For the unconstrained quadratic  
\(q(\vec{y})=\tfrac12\vec{y}^{\top}\vec{Q}\vec{y}+\vec{c}^{\top}\vec{y}\) we have
\(q(\vec{y})\ge \tfrac{\sigma_{\mathrm{Q}}}{2}\|\vec{y}\|_{2}^{2}-c_{0}\|\vec{y}\|_{2}\).
Since \(u(\vec{P},\bm{\pi})\ge -B\) for every feasible \(\vec{P}\),
any optimal \(\vec{y}^{\star}\) satisfies
\[
    B\geq \tfrac{\sigma_{\mathrm{Q}}}{2}\|\vec{y}^{\star}\|_{2}^{2}-c_{0}\|\vec{y}^{\star}\|_{2}
\]
which is a quadratic inequality in \(t:=\|\vec{y}^{\star}\|_{2}\).
Solving yields
\(t
  \le
  \bigl(c_{0}+\sqrt{c_{0}^{2}+2\sigma_{\mathrm{Q}}B}\bigr)\big/\sigma_{\mathrm{Q}}\).
Passing to \(\ell_{1}\)-norm by \(\|\vec{y}\|_{1}\le\sqrt{N}\|\vec{y}\|_{2}\) gives
\(
    \|\vec{y}^{\star}\|_{1}\le Y_{\max}.
\)
We then have
\[
          \|\vec{y}^{\star}\|_{1}
           \le 
          Y_{\max}
          :=
          \sqrt{N}\frac{\,
             c_0+\sqrt{c_0^2+2\sigma_{\mathrm{Q}}B}
          }{{\sigma_{Q}}}.
        \]
        Hence, the sample-wise constant is
        \[
          L_{(1)}(\vec{I})
           \le 
          \bigl(
            \|\vec{Q}\|_{1}\,\|\vec{P}\|_{1}\,Y_{\max}^{2}
            +c_0Y_{\max}
          \bigr)
          =:C(\varPi)'.
        \]
\end{proof}

\subsection{Discussion about neural network}
\label{discuss_theory}
\paragraph{Neural Network Architecture.}
We focus on fully connected architectures; extension to GNN networks is left for future work.

Following \cite{iwata2025learning}, we consider a feedforward neural network (NN) with $L$ hidden layers and architecture $[W_0, W_1, \ldots, W_L, W_{L+1}] \in \mathbb{N}^{L+2}$. The network maps a QP instance $\bm{\pi} = (\vec{Q}, \vec{c}, \vec{A}, \vec{b})$ to a projection matrix $\vec{P} \in \mathcal{P} \subset \mathbb{R}^{N \times K}$ via an encoder-decoder structure.

The input is encoded as a vector of length $W_0 = N^2 + N + MN + M$:
\[
\text{Enc} : \mathbb{R}^{N \times N} \times \mathbb{R}^N \times \mathbb{R}^{M \times N} \times \mathbb{R}^M \to \mathbb{R}^{W_0}.
\]
The output layer of size $W_{L+1} = NK$ is decoded into a matrix:
\[
\text{Dec} : \mathbb{R}^{NK} \to \mathbb{R}^{N \times K}.
\]

The NN consists of $L+1$ affine maps $T_\ell : \mathbb{R}^{W_{\ell-1}} \to \mathbb{R}^{W_\ell}$, where $T_{L+1}$ is linear. The model is parameterized by $\theta \in \mathbb{R}^W$, where $W$ is the total number of parameters. The mapping is written as:
\[
f^N_\theta(\bm{\pi}) = \text{Dec}\left(T_{L+1}\left(\sigma\left(T_L\left(\cdots \sigma\left(T_1(\text{Enc}(\bm{\pi}))\right) \cdots \right)\right)\right)\right),
\]
where $\sigma$ is the hidden-layer activation.

To ensure the output lies in $[-1,1]^{W_{L+1}}$, we apply a coordinate-wise squeezing function $\sigma': \mathbb{R} \to [0,1]$ to the output:  $\eta_i + (\tau_i - \eta_i)\sigma'(y_i)$. We use ReLU and clipped ReLU as activation functions for $\sigma$ and $\sigma'$:
\[
\sigma(x) = \max\{0, x\}, \quad \sigma'(x) = \min\{\max\{0, x\}, 1\}.
\]

To simplify the notation, let $U:=\sum_{\ell=1}^{L} W_{\ell}$ be the total number of hidden units and
recall that $W$ denotes the number of weights of the network.
Set $W_{0}=N^2+N+MN+M$ (length of the QP encoding) and $W_{L+1}=NK$ (length of the
vectorized projection matrix). Denote by $L$ the number of hidden layers. We use $W$ to denote the number of parameters of an NN.

Then \cite{iwata2025learning} provided the result that the pseudo-dimension~\cite{anthony2009neural} of the network defined above satisfies
$d_p =
\mathcal{O} \Bigl(WL\log(U+NK)\Bigr)$ at least for each dimension.
Bounding the covering number by the pseudo-dimension, we have $\log\mathcal{N}(\epsilon,\mathcal{F},D) =\mathcal{O}(NK d_p \log\frac{nB}{d_p\epsilon})$ from \cite{anthony2009neural}.

Setting $\epsilon=\mathcal{O}\!\bigl(1/(C(\varPi)\sqrt{D})\bigr)$ leads to the bound of Theorem~\ref{thm_generalization}. The generalization upper bound is of order
%\begin{align}
%    &\sup_{f\in\mathcal{F}}\Bigl|\,\frac1D\sum_{d=1}^{D} u(f(\pi_d),\pi_d) -\mathbb{E}_{\pi\sim\mathcal{D}}\bigl[u(f(\pi),\pi)\bigr]\Bigr|\notag \\
\begin{align}
\leq \mathcal{O}\left(\sqrt{\frac{NK d_p B^2\log (DB C(\varPi)/d_p)}{D}}\right)+\sqrt{\frac{2B^2\log\frac{2}{\delta}}{D}}.\notag
\end{align}
We can see that the generalization gap decreases as we increase $D$. This is a special case of the derived bound in Section~\ref{sec:theory}, where we assume a general parametric model with $\beta$-Lipshitz continuity.

\section{Data}
\label{app:data}

\paragraph{Regression} This dataset contains QP instances of
constrained linear regression~\cite{geweke1986exact},
\begin{align}
  \min_{\vec{x}}
  \parallel\bm{\beta}-\bm{\Phi}\vec{x}\parallel^{2}
  \quad
  \mathrm{s.t.}
  \quad
  \vec{A}'\vec{x}\leq\vec{b}',\quad\vec{x}\geq\vec{0},
\end{align}
where 
$\bm{\Phi}\in\mathbb{R}^{T\times N}$ is the design matrix of $T$ training data points for regression,
$\bm{\beta}\in\mathbb{R}^{T}$ is their targets,
$\vec{x}\in\mathbb{R}^{N}$ is linear coefficients, and
$\vec{A}'\in\mathbb{R}^{M\times N}$ and 
$\vec{b}'\in\mathbb{R}^{M}$ define the inequality constraints.

This problem can be rewritten in the standard form of
Eq.~\eqref{eq:qp} by defining
$\vec{Q}=2\bm{\Phi}^{\top}\bm{\Phi}\in\mathbb{R}^{N}\times \mathbb{R}^{N}$,
$\vec{c}=-2\vec{\Phi}^{\top}\vec{\beta}\in\mathbb{R}^{N}$,
$\vec{A}=[\vec{A}',-\vec{I}_{N}]\in\mathbb{R}^{M+N}$,
and $\vec{b}=[\vec{b}',\vec{0}]\in\mathbb{R}^{M+N}$.
To generate instances, we drew entries of $\bm{\Phi}$ and $\bm{\beta}$
independently from uniform distribution $\mathcal{U}(-1, 1)$,
and entries of $\vec{A}'$ and $\vec{b}'$ from $\mathcal{U}(0, 1)$,
where $\vec{b}'$ was scaled by $N$.  
We set the number of data points to $T = 2N$,
and the number of inequality constraints to $M = 50$.

\paragraph{Portfolio}
This dataset consists of portfolio optimization instances
that minimize the variance (i.e., risk)~\cite{best2000quadratic},
\begin{align}
  \min_{\vec{x}} \frac{1}{2}\vec{x}^{\top}\vec{Q}\vec{x}
  \quad
  \mathrm{s.t.}
  \quad\vec{1}^{\top}\vec{x}=1, 
  \quad\vec{x}\geq 0,
  \quad\bm{\mu}^{\top}\vec{x}\geq R,
\end{align}
where $\vec{x}\in\mathbb{R}^{N}$ is portfolio weights,
$\vec{Q}\in\mathbb{R}^{N\times N}$ is the covariance matrix between assets,
$\bm{\mu}\in\mathbb{R}^{N}$ is expected returns,
$R$ is the target return,
and $\vec{1}=[1,\dots,1]^{\top}$ is the all-ones vector.
This can be represented in the form of Eq.~(\ref{eq:qp_eq}) by setting
$\vec{A}_{\mathrm{E}}=\vec{1}$,
$\vec{b}_{\mathrm{E}}=1$,
$\vec{A}_{\mathrm{I}}=[-\vec{I}_{N},-\bm{\mu}]\in\mathbb{R}^{(N+1)\times N}$,
and
$\vec{b}_{\mathrm{I}}=[\vec{0},R]\in\mathbb{R}^{N+1}$.
We transformed this QP with linear inequality and equality constraints
to a QP with inequality constraints
using a feasible solution $\vec{x}_{0}=\frac{\vec{1}}{N}$
as described in Appendix~\ref{app:transform}.
The covariance matrix was computed by
$\vec{Q}=\vec{Q}_{0}^{\top}\vec{Q}_{0}+10^{-2}\vec{I}_{N}$,
where $\vec{Q}_{0}\in\mathbb{R}^{N\times N}$ was
randomly generated from the standard normal distribution.
We randomly generated $\bm{\mu}$ from $\mathcal{U}(-0.2,0.2)$,
and set the target by the average of the expected returns
$R=\frac{1}{N}\bm{\mu}^{\top}\vec{1}$.

\paragraph{Control} This dataset comprises quadratic optimal control problems
with linear dynamics and box constraints~\cite{petersen2006constrained},
\begin{align}
  &\min_{\{\vec{s}_{t}\vec{v}_{t}\}_{t=1}^{T}}
  \frac{1}{2}\sum_{t=1}^{T}\left(\parallel\vec{s}_{t}-\vec{s}^{*}\parallel^{2}+\mu\parallel\vec{v}_{t}\parallel^{2}\right),
  \nonumber\\
  &\mathrm{s.t.}
  \quad\vec{s}_{t+1}=\vec{s}_{t}+\vec{R}\vec{v}_{t}\quad
  (t=1,\dots,T-1),
  \quad
  \vec{s}_{1}=\tilde{\vec{s}},
  \nonumber\\
  &\underline{\vec{s}}\leq \vec{s}_{t}\leq \bar{\vec{s}},\quad
  \underline{\vec{v}}\leq \vec{v}_{t}\leq \bar{\vec{v}}\quad
  (t=1,\dots,T),
\end{align}
where $\vec{s}_{t}\in\mathbb{R}^{S}$ is the state at time $t$,
$\vec{v}_{t}\in\mathbb{R}^{V}$ is the control input,
$\vec{s}^{*}\in\mathbb{R}^{S}$ is the target state,
$\mu\in\mathbb{R}_{>0}$,
$\vec{s}_{t+1}=\vec{s}_{t}+\vec{R}\vec{v}_{t}$ defines the linear dynamics
with $\vec{R}\in\mathbb{R}^{S\times V}$,
$\tilde{\vec{s}}$ is the initial state, and
$\underline{\vec{s}}, \bar{\vec{s}}, \underline{\vec{v}}, \bar{\vec{v}}$ 
define the element-wise lower and upper bounds on state and control variables.
This problem include $(S+V)T$ variables,
$2(S+V)T$ inequality constraints,
and $ST$ equality constraints.
The linear inequality constraints
were eliminated using a feasible solution
$\vec{s}_{t}=\tilde{\vec{s}}$ and $\vec{v}_{t}=\bm{0}$ for $t=1,\dots,T$
as described in Appendix~\ref{app:transform}.
We set the number of time steps $T = 5$,
and both the state and control dimensions to $S = V = 50$,
yielding $N = 500$ total variables.  
We generated $\underline{\vec{s}}$ and
$\underline{\vec{v}}$ from $\mathcal{U}(-1,0)$,
$\bar{\vec{s}}$ and $\bar{\vec{v}}$
from $\mathcal{U}(0,1)$,
$\vec{s}^{*}$ and $\tilde{\vec{s}}$ from $\mathcal{U}(\underline{\vec{s}},\bar{\vec{s}})$,
$\mu$ from $\mathcal{U}(0,2)$,
and $\vec{R}$ from $\mathcal{U}(-1,1)$.

\section*{Acknowledgments}
FF was supported by JSPS KAKENHI Grant Number JP23K16948. FF was supported by JST, PRESTO Grant Number JPMJPR22C8, Japan.

\bibliographystyle{abbrv}
\bibliography{tnnls2025}

\end{document}